\documentclass[letterpaper,12pt,leqno]{article}
\usepackage{minimalpaper}
\usepackage{ragged2e}

\usepackage{pifont}
\newcommand{\cmark}{\ding{51}}%
\newcommand{\xmark}{\ding{55}}%
\hypersetup{pdftitle={GNN with Transformer Fusion}}

\begin{document}

\title{Graph Neural Networks with Transformer Fusion of Brain Connectivity Dynamics and  Tabular Data for  Forecasting  Future Tobacco Use}

    \author{Runzhi  Zhou    and Xi Luo
        \thanks{Runzhi Zhou  and Xi Luo are with  Department of Biostatistics and Data Science, The University of Texas Health Science Center at Houston, Houston, TX 77030 USA.  Corresponding author Xi Luo (xi.rossi.luo@gmail.com, xi.luo@uth.tmc.edu). Mail address:  1200 Pressler St, Houston, TX 77030 USA. }
        \thanks{The code implementation of our research will be made publicly available on GitHub upon the acceptance of our work for publication.}
    }
\date{}


\begin{titlepage}
    \maketitle
    \begin{center}
        \textbf{Abstract}
    \end{center}
    Integrating non-Euclidean brain imaging data with Euclidean tabular data, such as clinical and demographic information, poses a substantial challenge for medical imaging analysis, particularly in forecasting future outcomes. While machine learning and deep learning techniques have been applied successfully to cross-sectional classification and prediction tasks, effectively forecasting outcomes in longitudinal imaging studies remains challenging.
    To address this challenge, we introduce a time-aware graph neural network model with transformer fusion (GNN-TF). This model flexibly integrates both tabular data and dynamic brain connectivity data, leveraging the temporal order of these variables within a coherent framework. By incorporating non-Euclidean and Euclidean sources of information from a longitudinal resting-state fMRI dataset from the National Consortium on Alcohol and Neurodevelopment in Adolescence (NCANDA), the GNN-TF enables a comprehensive analysis that captures critical aspects of longitudinal imaging data.
    Comparative analyses against a variety of established machine learning and deep learning models demonstrate that GNN-TF outperforms these state-of-the-art methods, delivering superior predictive accuracy for predicting future tobacco usage.
    The end-to-end, time-aware transformer fusion structure of the proposed GNN-TF model successfully integrates multiple data modalities and leverages temporal dynamics, making it a valuable analytic tool for functional brain imaging studies focused on clinical outcome prediction.  

    \textbf{Keywords:}     dynamic functional connectivity,  graph neural networks, tobacco use, fusion, transformer, longitudinal fMRI, NCANDA.
\end{titlepage}

\section{Introduction}
\label{sec:introduction}
Globally, tobacco use is acknowledged as the foremost preventable cause of death and disease, with the number of deaths projected to rise from 5 million to 8 million annually by 2030 \citep{warren2009change}. The widespread prevalence of smoking among children and adolescents is a major public health concern. A seminal study conducted in 1995 in the United States revealed that 20\% of premature deaths related to tobacco among individuals under 17 could have been prevented if they had not started smoking \citep{epstein2000model}. Typically beginning in adolescence, smoking behavior is prone to persist into adulthood, significantly contributing to the escalating rates of tobacco-related diseases. These deeply ingrained behaviors are challenging to alter, highlighting the critical need for early detection and targeted prevention programs aimed at high-risk adolescent smokers as vital public health initiatives.

In the present study, our primary focus lies in harnessing longitudinal data obtained from Functional Magnetic Resonance Imaging (fMRI) in conjunction with demographic data from the National Consortium on Alcohol and Neurodevelopment in Adolescence (NCANDA) study \citep{brown2015national}. Our attention is particularly directed towards resting-state fMRI, a pivotal technique in exploring brain functional connectivity \citep{biswal1995functional}. This methodology has been crucial in uncovering links between brain activity and a range of cognitive and behavioral outcomes \citep{van2010exploring,zhang2021have}. Importantly, it is increasingly acknowledged that alterations in functional connectivity attributable to disorders and diseases broadly affect neural networks, extending beyond localized brain regions \citep{zhang2021have}.  In a recent review \citep{boer2022brain}, structural changes observed via MRI have been shown to predict future tobacco and other substance use in adolescents, providing evidence of specific brain structure vulnerabilities for future use. However, evidence from longitudinal functional MRI studies remains limited.
In adolescents, resting-state functional connectivity is integral to neurodevelopment \citep{ernst2015fmri} and has the potential to identify vulnerable connectivity patterns and populations.


To gain a more comprehensive understanding of functional connectivity (FC)   networks and their associations with diverse outcomes \citep{lurie2020questions}, researchers have increasingly shifted their focus from static FC to time-varying FC, leveraging resting-state fMRI data. This shift represents a significant evolution in the field, underscored by the rapid growth and burgeoning interest in time-varying FC \citep{lurie2020questions}. Predominantly, these studies utilize the sliding window technique to analyze pairwise correlations, thereby establishing it as the de facto standard method \citep{lurie2020questions}. However, a considerable limitation of most time-varying FC methods is their inadequate focus on coherently summarizing temporally ordered FC, particularly in relation to predicting future disease outcomes. The critical role of incorporating the temporal sequence in FC matrices has been recognized and previously explored in research \citep{yaesoubi2015dynamic}.

Recent advancements in machine learning and deep learning have profoundly influenced the development of innovative methods for analyzing  functional connectivity from resting-state fMRI data \citep{khosla2019machine}. These methods have proven instrumental in predicting social anxiety disorders \citep{Mansson2015PredictingLO} and diagnosing neurological conditions \citep{Devika2021AMLA}. Notably, the application of Convolutional Neural Networks (CNN) in Alzheimer’s Disease analysis \citep{Sun2022TemporalAS} and the utilization of Recurrent Neural Network (RNN) models, such as Long Short-Term Memory (LSTM) models \citep{hochreiter1997long},  with the NCANDA dataset for studying adolescents with alcohol use disorders \citep{Ouyang2020LongitudinalPA} highlight the versatility of these advancements. In network neuroscience, Graph Neural Networks (GNN) have emerged as a crucial tool for analyzing non-Euclidean graph data, such as functional connectivity \citep{bessadok2022graph}. This contrasts with traditional tabular variables like sex and age, which are typically represented in Euclidean space. Applications of GNNs to functional connectivity data have included cross-sectional predictions of age, sex \citep{gadgil2020spatio, Li2022JointGC}, and various diseases \citep{ktena2018metric, wang2021graph}. Prior applications of GNN and LSTM have focused on examining dynamic brain networks for disease prediction without integrating tabular data \citep{Cao2022ModelingTD, Liu2023BrainTGLAD}. These efforts have either targeted outcomes concurrent with the fMRI data or overlooked the integration of tabular data in their models. However, recent advancements in sequence modeling, particularly with transformers \citep{vaswani2017attention}, such as large language models \citep{chang2023survey,thirunavukarasu2023large}, have opened new avenues for enhancing predictions based on sequences.

Simultaneously, a myriad of machine learning and deep learning models have been developed specifically for Euclidean tabular predictors. In the context of fMRI research, it is common practice to collect various subject characteristics, including demographic and clinical variables, in addition to outcome measures and fMRI data. These variables are important biological variables of interest and should be integrated into analytical models. Intrinsically, brain networks are non-Euclidean in nature. To integrate both types of data, traditional methods attempt to simplify this complexity by undertaking feature engineering, such as the extraction of key network attributes via graph-theoretical analysis \citep{bullmore2009complex}. These techniques transform non-Euclidean graph data into Euclidean space, facilitating the application of numerous Euclidean-based methods. However, the potential information loss resulting from feature engineering is a significant concern. Additionally, the complex, nonlinear relationships between graph data and tabular data might be inadequately addressed in these feature engineering attempts, prompting questions about the thoroughness of such methodologies.

Despite these advancements, a significant gap persists in effectively integrating time-varying functional connectivity with tabular data in a holistic, end-to-end approach, especially respecting the temporal order. To address this, our study introduces the Graph Neural Network Transformer Fusion (GNN-TF), a novel model specifically designed to concurrently analyze brain dynamic connectivity and tabular data within a unified neural network framework. This approach aims to harness the potential of existing methodologies while uniquely accounting for the temporal sequence of events, thereby offering a new perspective in understanding  brain dynamics. In our study, there is a natural temporal order among the tabular data, dynamic connectivities, and future tobacco use outcomes, which our model inherently respects. This logical, time-aware integration enhances prediction performance.

We validated our GNN-TF approach using a longitudinal fMRI dataset from the  NCANDA study, focusing on the challenging task of forecasting future tobacco use. Furthermore, we conducted extensive comparisons with other machine learning and deep learning models. The contributions of this paper are as follows:

\begin{enumerate}
    \item  Introduction of an end-to-end, temporally-aware Transformer Fusion structure for integrating multiple non-Euclidean and Euclidean data sources within a unified framework.
    \item Extensive comparison with various machine learning and deep learning models in evaluation studies.
    \item Integration of our model with state-of-the-art pretrained transformer models using transfer learning  to enhance its predictive capabilities.
    \item Application and comparison of various  GNN backbone structures \citep{Kipf2016SemisupervisedCW,Velickovic2017GraphAN,Brody2021HowAA,Xu2018HowPA,Gravina2022AntiSymmetricDA,Gasteiger2018PredictTP,Monti2017GeometricDL}  within our framework, offering a comprehensive assessment of their performance.
    \item Evaluation of the model's efficacy in forecasting future tobacco use using a public, medium-sized longitudinal fMRI dataset from NCANDA.
\end{enumerate}

\section{Materials and Methods}
\subsection{Dataset and Preprocessing}
This study utilized the   curated dataset from the NCANDA study \citep{brown2015national}, a multi-site longitudinal neuroimaging study. The NCANDA study investigates adolescents at diverse stages of development using an accelerated longitudinal design. The study tracks a cohort of youths aged 12–21 years (49\% male, assigned at birth; 64\% white; over 50\% at risk of heavy drinking), from baseline and annually for up to nine years.

This study included resting-state fMRI (rs-fMRI) data and key biological variables, all collected at baseline (year 0). The pivotal biological variables were age and sex assigned at birth. In addition, self-reported tobacco use was monitored in baseline and follow-up visits.  Participants diagnosed with any substance use disorder according to DSM IV (1994) at baseline were excluded. To investigate the onset of tobacco use in tobacco-naïve individuals, subjects who reported tobacco use at baseline were also excluded \citep{sharapova2020age}. The outcome variable, future tobacco use,  was identified as any use reported during follow-up visits in years 1-3, as available in the current data release at the time of this analysis. The fMRI images were processed using the C-PAC pipeline \citep{craddock2013towards}. Preprocessing steps included optimized brain extraction \citep{lutkenhoff2014optimized}, discarding the first five scans for stabilization, slice timing correction, nonlinear registration to MNI space, ICA-AROMA denoising \citep{pruim2015ica}, linear detrending, motion correction, and CompCor \citep{behzadi2007component}. ROI-based average time series were extracted for each participant using either the Power 264 atlas \citep{power2011functional} or a customized Harvard-Oxford atlas \citep{zhao2021principal}.   Subjects lacking   follow-up data or with fMRI data not meeting quality standards were excluded, resulting in $N=522$ subjects contributing to multivariate time series matrices.

\subsection{Method Overview}
\begin{figure*}[!t]
    \centering
    \includegraphics[width=\textwidth]{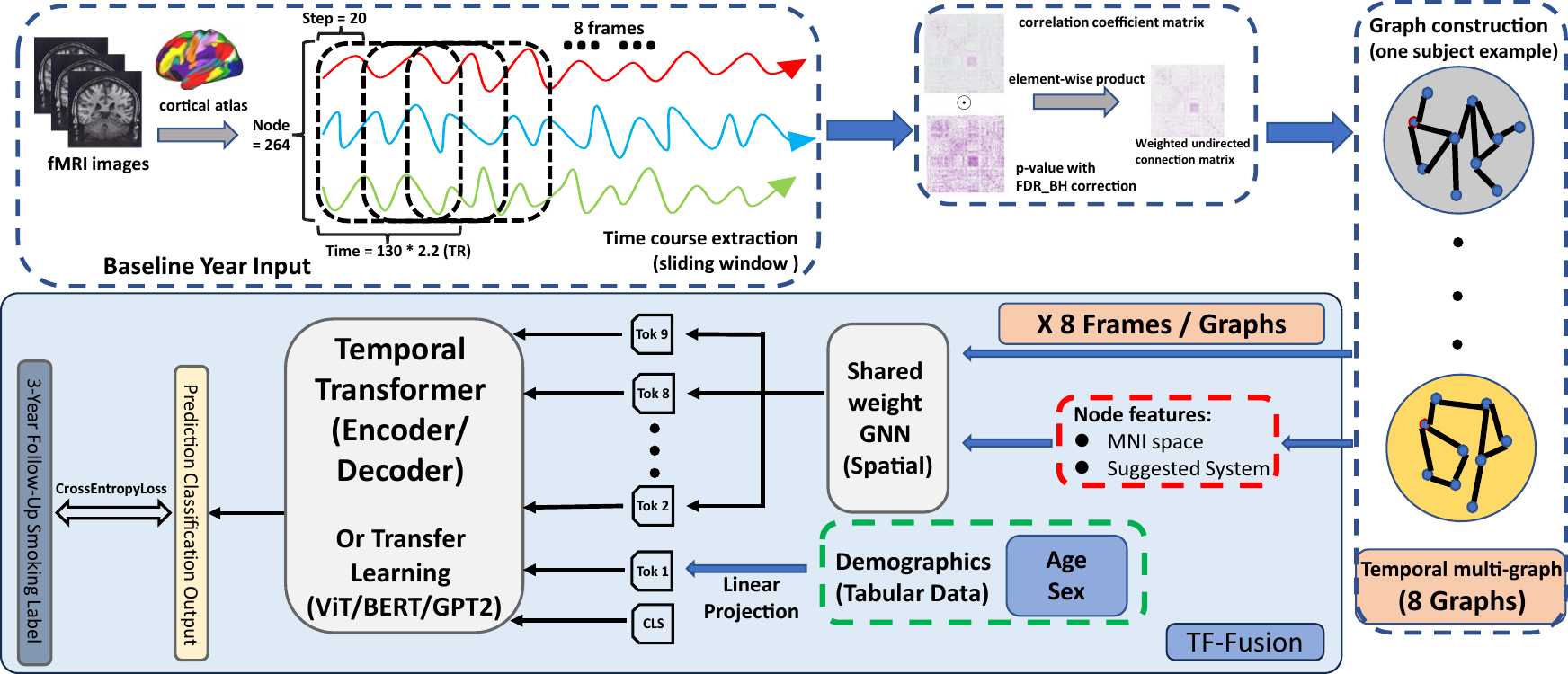}
    \caption{The proposed GNN-TF network structure for fMRI imaging and structured data in classification tasks. A classification token ``cls'' is added to the beginning of the sequence in all transformer models as a prompt, except for GPT2. In GPT2, following the guidelines in the OpenAI GPT2 documentation from the Python package ``transformers'', ``cls'' is placed at the end of the sequence, with sex and age features being projected to the penultimate position.}
    \label{fig:workflow}
\end{figure*}

Graph neural networks (GNNs) have become instrumental in analyzing network data and forecasting outcomes.   In our enhancement to the GNN framework, we integrated tabular variables, including both raw measures and hand-crafted features, into a unified GNN model. Additionally, we incorporated a transformer network structure for handling temporal brain connectivity dynamics using transfer learning. The brain connectivity networks were constructed using   Pearson's correlations between brain regions, based on resting-state fMRI time series from  popular brain atlases \citep{power2011functional,zhao2021principal}. Multiple testing correction was employed to foster sparse connections and reduce false positives. Our model utilized raw measures derived from subject-level variables such as age and sex. The hand-crafted features comprised node-level attributes related to the biological characteristics (e.g., location and functional segmentation) of brain regions.
We chose these tabular variables due to data availability and our scientific objectives, though our methodological framework is adaptable to incorporate other types of tabular variables. Our GNN-TF network structure is illustrated in Fig.~\ref{fig:workflow}.

\subsection{Functional Connectivity Construction}\label{subsec:fc_construction}
The basic Pearson's correlation matrices can be  interpreted as fully connected, undirected weighted networks. These networks differ from typical  GNN input data due to their fully connected nature. Historically, thresholding the correlation matrices has been a popular approach before applying GNN models, with the threshold level often treated as a critical hyperparameter \citep{wang2023brain}. In our study, we employed a data-driven, multiple testing correction approach for thresholding correlation matrices, following a multi-step procedure.

Initially, Pearson's correlation matrices $C^{(is)}$, for $i=1,2,\dotsc, N$, $s = 1, 2, \dotsc, 8$, were computed to quantify brain connectivity from participant $i$ and sliding time window $s$. Here we utilized a sliding window approach to calculate dynamic functional connectivity to be described momentarily. Following this, the Benjamini/Hochberg False Discovery Rate Correction (FDR) was applied to the matrix of p-values.   Subsequently, binary adjacency matrices were generated. In these matrices, FDR corrected p-values of 0.05 or less were denoted as 1, indicating a significant connection, while p-values above 0.05 were marked as 0, representing the absence of a significant connection. This process resulted in a binary matrix $B^{(is)}$ for each $i$ and $s$.

The final step involved computing weighted graphs $A^{(is)} = B^{(is)} \odot C^{(is)}$, where element-wise Hadamard multiplication $\odot$ was performed between the binary adjacency matrix and the correlation coefficients matrix for each $i$ and $s$. For the sake of simplicity, we will omit the superscripts $i$ and $s$ in the notation for $A^{(is)}$ in the subsequent discussion. 

\subsection{Graph Neural Network}
In our study, we chose to focus on the Graph Convolutional Network (GCN) \citep{Kipf2016SemisupervisedCW} structure for graph data, motivated by its computational efficiency and its performance, which can be comparable to or sometimes even superior to other models \citep{wu2021performance}. While the substitution of GCN with other  GNN backbone architectures \citep{Velickovic2017GraphAN,Brody2021HowAA,Xu2018HowPA,Gravina2022AntiSymmetricDA,Gasteiger2018PredictTP,Monti2017GeometricDL}  is straightforward, our description and analysis primarily centered on the GCN structure.
Our model consisted of two convolutional layers, with the operation at the $l$th layer described as follows:
\begin{align}
    H^{(l)} =  f(H^{(l-1)}, A) = \sigma( {D}^{-\tfrac{1}{2}} {A}  {D}^{-\tfrac{1}{2}} H^{(l-1)} W^{(l-1)}).
\end{align}
Here, $D$ is a diagonal matrix with entries $D_{ii} =  \sum_j \mathbf{1}\{A_{ij} \ne 0 \}$, where $\mathbf{1}\{ \cdot \}$ is the indicator function. $H^{(l)}$ represents the output from the $l$th layer, and $W^{(l)}$ denotes the learnable parameter matrix at the $l$th layer. $\sigma(\cdot)$ is the activation function. We set $W^{(l)}$ with dimensions such that the outputs $H^{(l)}$, for $l = 1, 2$, both had a dimension of 128. The Scaled Exponential Linear Unit (SELU) \citep{Klambauer2017SelfnormalizingNN} was used as the activation function throughout the network.

In our GCN, the nodes corresponded to brain regions. We set the input node features $H^{(0)}$ to the MNI space coordinates in the $x$, $y$, and $z$ directions, along with the one-hot encoding of suggested system classifications \citep{power2011functional}. A fully connected layer with an output dimension of 128 was included as the final layer to project the convolutional output into a space relevant to our forecasting task.

\subsection{Transformer Fusion}\label{subsubsec:tf}
To capture the dynamics of the brain, we employed a sliding window strategy, a common approach in this field \citep{hutchison2013dynamic}. Considering computational time, we opted for $8$ sliding windows, each with a width of $130$ TR and a step size of $20$ TR (where TR = 2.2 seconds), aligning with recommended window sizes in \citep{Savva2019AssessmentOD}. This window size was chosen to capture variation over a reasonable time span, essential for our goal of forecasting future smoking outcomes in 3 years. Moreover, this sample size in each sliding window provides sufficient statistical power for our network construction, as detailed in Section~\ref{subsec:fc_construction}.

Our forecasting problem combines dynamic brain network features and tabular variables into a sequence classification framework. We framed this challenge analogously to similar problems encountered in computer vision and natural language processing, and  utilized transformer models \citep{vaswani2017attention} to integrate dynamic GNN features with tabular variables. Various transformer models, including pretrained ones, were incorporated into our framework. We experimented with training the standard transformer model \citep{vaswani2017attention} from scratch, as well as employing popular pretrained transformers for other tasks, including ViT \citep{Dosovitskiy2020AnII}, BERT \citep{Devlin2018BertPO}, and GPT-2 \citep{Radford2018ImprovingLU}. The standard transformer model in our study had three layers, each with four-head attention. The input sequence's first position was set to the projected embedding of tabular variables (sex and age), followed by GNN features in their temporal order. This projection was achieved using a fully connected layer on batch-normalized input with SELU activation, ensuring the embedding dimension of 128 matched the GNN output. For transfer-learning tasks, a projection layer was added to align dimensions with pretrained temporal transformer models. An alternate fusion strategy, termed 'late fusion', diverges by not integrating variables directly into the transformer. Instead, they are fused outside the transformer by concatenating the transformer output with the covariate embedding before the prediction layer, as depicted in Fig.~\ref{fig:late_tf}. However, late fusion does not respect the temporal order of the tabular and imaging data. This may not provide the optimal strategy, nor does it enhance interpretability, which our proposed approach achieves through temporal awareness.
We will discuss results from this fusion approach and other state-of-the-art methods in ablation studies in Section~\ref{subsubsec:Additional}.

Finally, we employed the negative log-likelihood loss function for binary outcomes to evaluate the transformer predictions. This involved using three fully connected layer projections followed by a log-softmax layer, set against the future smoking status.

\section{Experiments and Results}
\subsection{Experimental Setup}

In our experiment, we adopted a stratified five-fold cross-validation strategy with inner-fold ensembling to ensure robust evaluation. In the outer loop, 20\% of the subjects were withheld as the testing set. Within the remaining training data, we employed an inner stratified five-fold cross-validation to train the models and determine early stopping. To mitigate the potential variance associated with GNN training, the final performance for each outer fold was computed by averaging the predictions of the five models trained in the inner loop.

To ensure a fair comparison, we adopted a unified set of hyperparameters, given that the trainable components (GNN encoders and downstream linear layers) were structurally similar across all models. For the training process, Adam was used as the optimizer with a learning rate of $10^{-4}$ and a batch size of $32$. The hidden layer dimension was set to $128$ to balance model capacity and overfitting risks. To further prevent overfitting, early stopping was implemented, guided by the Area Under the Curve (AUC) observed on the validation set.


Depending on the transformers used, our model variations were termed GNN-TF, GNN-TF-V, GNN-TF-B, and GNN-TF-G, corresponding to transformers trained from scratch, ViT, BERT, and GPT-2, respectively. We compared these models with the state-of-the-art dynamic GNN model, GC-LSTM \citep{Chen2022GCLSTMG}. GC-LSTM aggregates graph data at each temporal snapshot using GCN, and LSTM is used to learn temporal dynamics. This structure has been shown to be superior to several other proposals. The existing GC-LSTM does not incorporate tabular variables. For a fair comparison, we applied a late fusion strategy (similar to Fig.~\ref{fig:late_tf}) by concatenating the projected variables with the GC-LSTM output before feeding it into the final prediction layer. This modified version is termed GC-LSTM-F. We also compared our models with classical machine learning models, such as logistic regression (LR) and random forests (RF), as these are typically developed for non-sequence data. This comparison is detailed in our ablation studies in Section~\ref{subsubsec:Additional}.

\subsection{Performance Evaluation}
The area under the receiver operating characteristic curve (AUC) and the Precision-Recall AUC (PRAUC) were utilized to evaluate the classification performance of our models. These metrics are advantageous as they do not rely on the tuning of threshold parameters, unlike accuracy. They provide an overall assessment of performance without considering the trade-off between false positives and negatives. The chance value for AUC is $0.5$, while the chance level for PRAUC also varies depending on the prevalence. To ensure the robustness of our comparisons, each method was repeated five times, and the average AUC and PRAUC values across these repetitions were reported.

As summarized in Table~\ref{table:model_performance}, our comparative analysis of model performance highlighted the superior capabilities of the GNN-TF models across all variants. These models showed a notable enhancement in prediction performance compared to the state-of-the-art GC-LSTM models. In particular, the GNN-TF variants consistently outperformed the GC-LSTM models in both AUC and PRAUC metrics, underscoring their effectiveness in forecasting tasks. The performance remained similar across different brain atlases, indicating that our approach is robust against varying atlas choices. The enhanced results achieved by the GNN-TF models can be largely attributed to their ability to effectively integrate and learn from complex dynamic features, while simultaneously considering tabular covariates.

\begin{table}[ht]
    \centering
    \caption{Comparison of Prediction Performance}
    \label{table:model_performance}
    \begin{tabular}{lccccr}
        \toprule
                       & \multicolumn{2}{c}{\textbf{Power264}} & \multicolumn{2}{c}{\textbf{HO}} & \textbf{\# of Trainable}                                        \\
        \cmidrule(r){2-3} \cmidrule(r){4-5}
        \textbf{Model} & \textbf{AUC}                          & \textbf{PRAUC}                  & \textbf{AUC}             & \textbf{PRAUC} & \textbf{Parameters} \\
        \midrule
        GCLSTM         & 0.520                                 & 0.171                           & 0.447                    & 0.159          & 855,554             \\
        GCLSTM-F       & 0.658                                 & 0.263                           & 0.676                    & 0.256          & 872,326             \\
        GNN-TF         & 0.697                                 & 0.267                           & \textbf{0.710}           & {0.268}        & 1,917,190           \\
        GNN-TF-V       & 0.676                                 & 0.244                           & 0.702                    & 0.257          & 390,662             \\
        GNN-TF-B       & 0.665                                 & \textbf{0.274}                  & 0.695                    & \textbf{0.281} & 390,662             \\
        GNN-TF-G       & \textbf{0.700}                        & 0.271                           & 0.698                    & 0.250          & 491,782             \\
        \bottomrule
    \end{tabular}
    \par\smallskip
    \small
    \justifying
    {Average AUC and PRAUC metrics  using the Power 264 (Power264) atlas \citep{power2011functional} and the customized Harvard-Oxford (HO) atlas \citep{zhao2021principal},  and their trainable  parameters. The best performed for each category is highlighted in bold.}
    \par
\end{table}

\subsection{Ablation Studies}\label{subsubsec:Additional}
\begin{figure}[!t]
    \centerline{\includegraphics[width=\columnwidth]{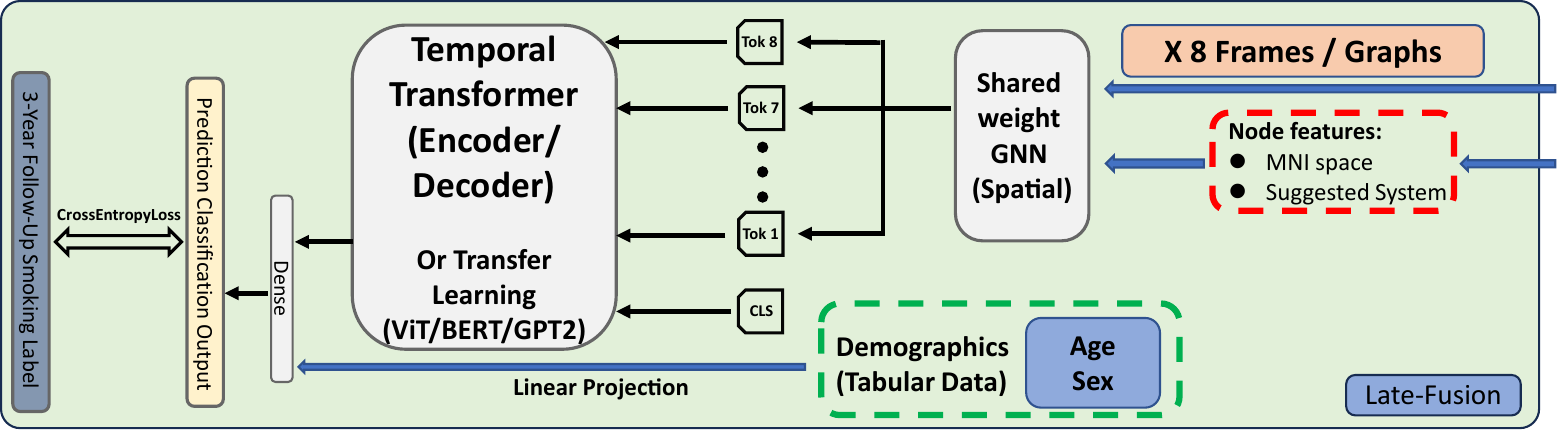}}
    \caption{In ablation studies, an alternative late-fusion strategy (inside the light green box above)  replace the TF-fusion module (light blue) in Fig.~\ref{fig:workflow}.}
    \label{fig:late_tf}
\end{figure}

We conducted a series of ablation studies to evaluate and compare models under various conditions, using the same experimental setup and metrics previously described. These studies involved multiple modifications to test the robustness and flexibility of our models. Modifications included replacing the base predictive models with alternative machine learning and GNN architectures, excluding covariates, omitting functional connectivity data, abandoning the sliding window method for calculating dynamic connectivity, and excluding transformer fusion as detailed in Section~\ref{subsubsec:tf}. The machine learning predictive models we evaluated included Random Forests (RF) and Logistic Regression (LR). Additionally, we explored a variety of  GNN backbone architectures. These comprised the Graph Attention Network (GAT) \citep{Velickovic2017GraphAN}, GATv2 \citep{Brody2021HowAA}, the Graph Isomorphism Network (GIN) \citep{Xu2018HowPA}, the Anti-Symmetric Graph Convolutional Layer (AntiSymmetricConv) \citep{Gravina2022AntiSymmetricDA}, Approximate Personalized Propagation (APPNP) \citep{Gasteiger2018PredictTP}, and the Gaussian Mixture Model Convolutional Layer (GMMConv) \citep{Monti2017GeometricDL}.
Beyond our transformer fusion (TF) approach, we also examined models using an alternative fusion strategy known as late fusion described in Fig.~\ref{fig:late_tf}.
We summarized these findings in Table~\ref{tab:Additional}. 

The results  revealed that one of our proposed models, GNN-TF, consistently outperformed almost all other models in the evaluation. Among the total of 56 metrics assessed, only five instances were noted where other models achieved higher values than our proposed GNN-TF model. It's important to highlight that none of these models surpassed the GNN-TF model in both AUC and PR AUC simultaneously. Upon a closer examination of these five metrics, we observed that the differences in performance were relatively minimal. Moreover, when comparing against a more advanced variant of our model, GNN-TF-G, the number of metrics where other models performed slightly better reduced to only three out of the total 56.

\begin{table}[ht]
    \centering
    \caption{Additional Comparisions of Performance}
    \label{tab:Additional}
    \begin{tabular}{lccccc}
        \toprule
        \textbf{Backbone} & \textbf{Tabular} & \textbf{Connectivity} & \textbf{Dynamic} & \textbf{TF} & \textbf{AUC, Accuracy}      \\
        \midrule
        GCN, Transformer  & \cmark           & \cmark                & \cmark           & \cmark      & 0.697, 0.267 (our baseline) \\
        RF                & \xmark           & \cmark                & \xmark           & \xmark      & 0.551, 0.188                \\
        RF                & \cmark           & \xmark                & \xmark           & \xmark      & 0.600, 0.206                \\
        RF                & \cmark           & \cmark                & \xmark           & \xmark      & 0.552, 0.185                \\
        LR                & \xmark           & \cmark                & \xmark           & \xmark      & 0.520, 0.183                \\
        LR                & \cmark           & \xmark                & \xmark           & \xmark      & 0.672, \textbf{0.268}       \\
        LR                & \cmark           & \cmark                & \xmark           & \xmark      & 0.582, 0.207                \\
        GCN               & \xmark           & \cmark                & \xmark           & \xmark      & 0.530, 0.176                \\
        GCN               & \cmark           & \cmark                & \xmark           & \xmark      & 0.674, 0.254                \\
        GAT               & \xmark           & \cmark                & \xmark           & \xmark      & 0.563, 0.195                \\
        GAT               & \cmark           & \cmark                & \xmark           & \xmark      & 0.673, 0.260                \\
        GAT2              & \xmark           & \cmark                & \xmark           & \xmark      & 0.535, 0.179                \\
        GAT2              & \cmark           & \cmark                & \xmark           & \xmark      & 0.675, 0.260                \\
        GIN               & \xmark           & \cmark                & \xmark           & \xmark      & 0.507, 0.169                \\
        GIN               & \cmark           & \cmark                & \xmark           & \xmark      & 0.620, 0.243                \\
        AntiSymmetric     & \xmark           & \cmark                & \xmark           & \xmark      & 0.469, 0.152                \\
        AntiSymmetric     & \cmark           & \cmark                & \xmark           & \xmark      & \textbf{0.701}, 0.256       \\
        APPNP             & \xmark           & \cmark                & \xmark           & \xmark      & 0.501, 0.178                \\
        APPNP             & \cmark           & \cmark                & \xmark           & \xmark      & 0.682, 0.263                \\
        GMM               & \xmark           & \cmark                & \xmark           & \xmark      & 0.568, 0.208                \\
        GMM               & \cmark           & \cmark                & \xmark           & \xmark      & 0.670, \textbf{0.276}       \\
        GCN, Transformer  & \xmark           & \cmark                & \cmark           & \xmark      & 0.502, 0.163                \\
        GCN, Transformer  & \cmark           & \cmark                & \cmark           & \xmark      & 0.676, 0.259                \\
        GCN, ViT          & \xmark           & \cmark                & \cmark           & \xmark      & 0.522, 0.178                \\
        GCN, ViT          & \cmark           & \cmark                & \cmark           & \xmark      & 0.668, \textbf{0.276}       \\
        GCN, BERT         & \xmark           & \cmark                & \cmark           & \xmark      & 0.522, 0.180                \\
        GCN, BERT         & \cmark           & \cmark                & \cmark           & \xmark      & 0.674, 0.264                \\
        GCN, GPT2         & \xmark           & \cmark                & \cmark           & \xmark      & 0.484, 0.162                \\
        GCN, GPT2         & \cmark           & \cmark                & \cmark           & \xmark      & 0.672, \textbf{0.269}       \\
        \bottomrule
    \end{tabular}
    \par\bigskip
    \small \justifying
    {The performance changes in AUC and PRAUC, respectively,  were assessed by comparing wtih GNN-TF, the reference model on the first line. The late fusion strategy  was  used when all options except when TF  were checked.  Higher values in either AUC or PRAUC than the reference are highlighted in bold. TF: our transformer fusion approach. Only our baseline model, GNN-TF, is listed here for comparison, without the improvements from pretraining. Our other models may provide improved performance; see Table~\ref{table:model_performance}.
    }
    \par
\end{table}

\section{Discussion}

\subsection{General Discussion}
We introduced the use of  GNN with transformer fusion as a novel method to integrate dynamic brain networks and tabular data. Traditional statistical methods abound for assessing the association between tabular covariates and various outcomes, such as substance use and mental health conditions. However, these methods often fall short in handling the dynamic network data structure that GNNs are adept at modeling. Our approach bridges this methodological gap by utilizing deep learning models capable of incorporating both data types. This integration is particularly beneficial for functional imaging studies where both types of data are relevant to the outcomes.

Our method demonstrated superior performance compared to state-of-the-art alternatives. Additionally, its simplicity and flexibility facilitate the easy integration of pretrained transformer models. It’s important to note that the fields of GNN and deep learning are rapidly evolving, presenting other potential approaches that we did not explore. These opportunities are left for future research. Overall, our study underscores the importance of effectively fusing both brain dynamics and structural covariates in functional imaging research.

\subsection{Contributions of  Node Features and Connectivities}
To delve deeper into understanding the influential node features and connections that predicted the outcomes, we utilized GNNExplainer \citep{Ying2019GnnexplainerGE}, an optimization-based tool. For illustrative purposes, we based our analysis on the Power 264 atlas using our GNN-TF model. Due to computational constraints associated with GNNExplainer, a random subset of 32 subjects was selected for the analysis. We then computed the mean and 95\% confidence intervals ($\text{mean} \pm 1.96 \text{se}$) of the GNNExplainer results across these subjects.

\begin{figure}[!t]
    \centerline{\includegraphics[width=\columnwidth]{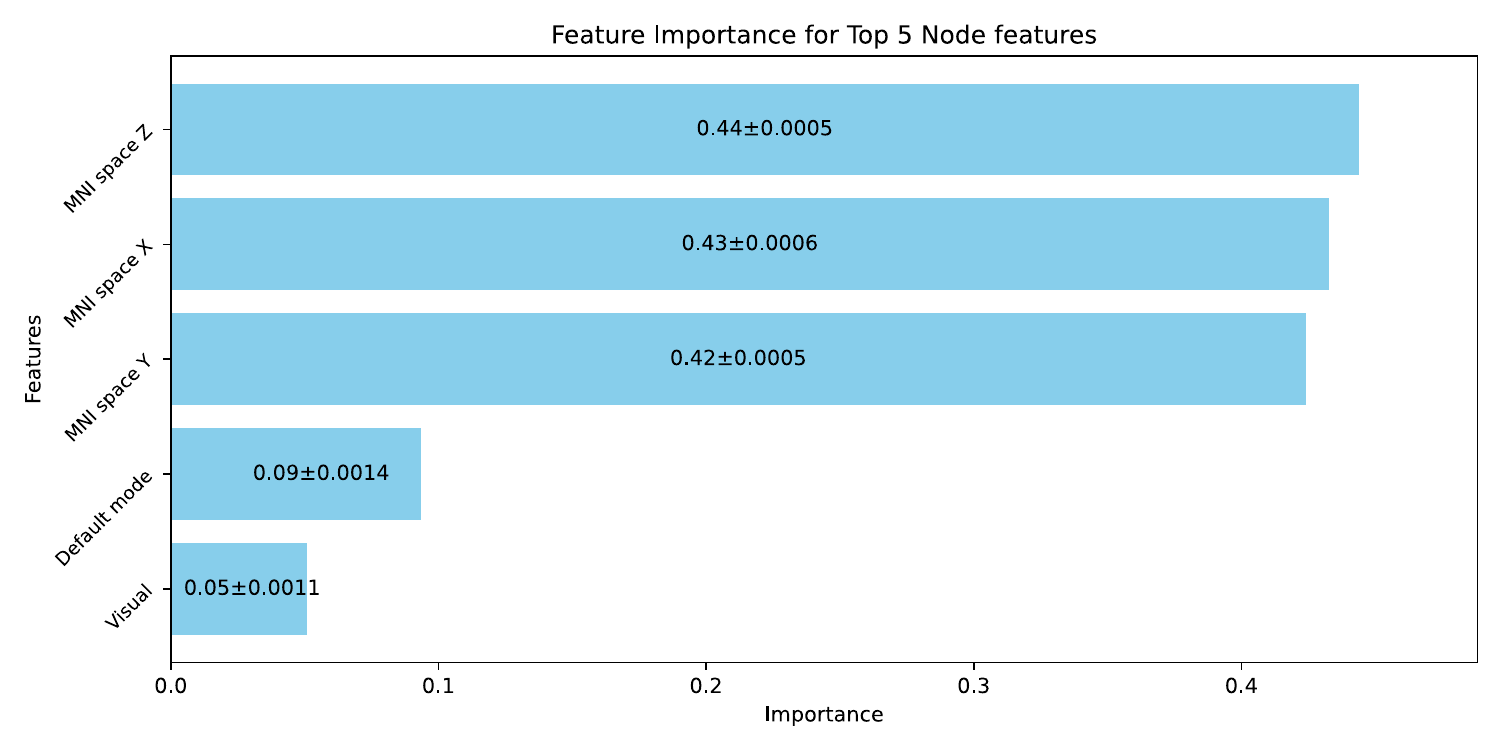}}
    \caption{Top five node features ranked by GNNExplainer. Feature importance values range from 0 (no impact on prediction) to 1 (the most impactful)}
    \label{fig:node}
\end{figure}

Fig.~\ref{fig:node} presented the top 5 node features identified by GNNExplainer. The findings highlighted the MNI space locations of the brain regions as the top three significant features in predicting outcomes, with importance values ranging decently between $0.42$ and $0.44$ (the best possible value being $1$). Additionally, regions associated with the default mode and visual systems also contributed to the prediction performance, albeit with a lesser magnitude of importance. These results indicated that structural information from specific brain regions, combined with functional brain connectivity graphs in our model, played a pivotal role in forecasting future tobacco use.

\begin{figure}[!t]
    \centerline{\includegraphics[width=\columnwidth]{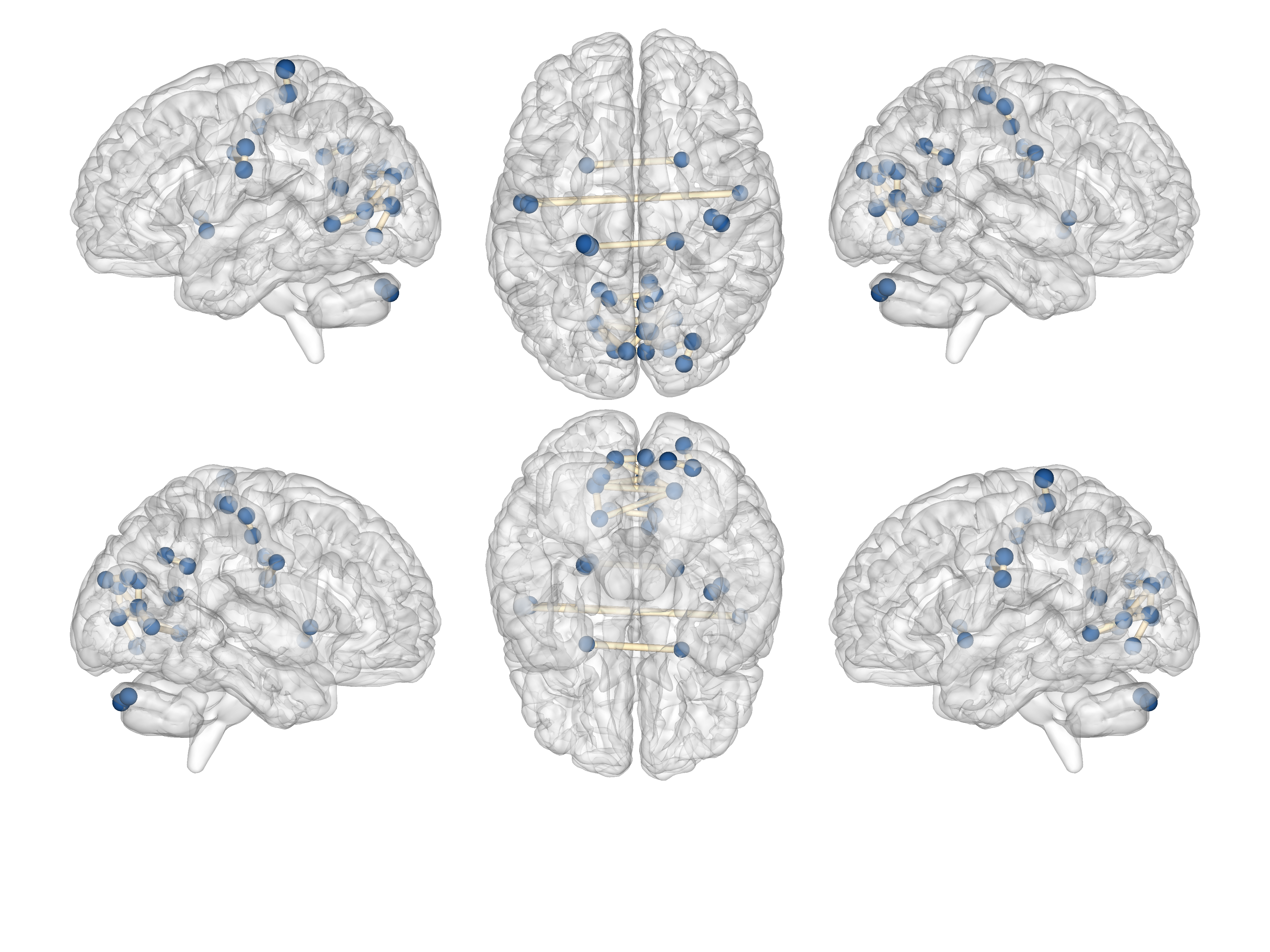}}
    \caption{Top 25 brain connections ranked by GNNExplainer.  The brain connections are shown in yellow, and the associated brain nodes are shwon in blue.}
    \label{fig:edge}
\end{figure}

Fig.~\ref{fig:edge} displayed the top 25 brain connections identified in our study.\footnote{Complete coordinates of these regions and their connections were provided in figshare  at blinded url 
} Notably, the majority of these connections were inter-hemispheric, linking almost symmetric regions across the two hemispheres of the brain. The brain regions primarily involved in these connections were predominantly located within the visual cortex and motor cortex. This pattern underscored the significance of these areas and their interconnectedness in the context of our study's focus.

\subsection{Limitations and Future Works}

Our study primarily utilized the NCANDA dataset, a longitudinal fMRI dataset of medium sample size. This unique design and sample size enabled the development of GNN fusion models with a reasonable number of parameters, allowing us to examine forecasting performance in light of the natural temporal order. However, it's noteworthy that the sample size in this study is larger than that typically found in single-site fMRI studies. In such cases, training deep learning models from scratch often risks overfitting. To counter this, we leveraged pretrained transformer models through transfer learning, effectively reducing the number of parameters. This approach may help mitigate overfitting issues, and its applicability to other datasets with varying sample sizes warrants further evaluation.

In our analyses, we focused on sex and age as covariates due to their biological significance and accurate ascertainment. While there are other potential covariates, such as those derived from behavioral questionnaires, that could be associated with the outcome, their inclusion remains to be explored. Incorporating additional covariates into our framework is straightforward, but it's yet to be determined if this would improve performance or if high-dimensional covariates might introduce overfitting problems.

Our choice of using sliding windows with the current time span was driven by our interest in predicting outcomes over a three-year period. Depending on the specific outcome of interest, these choices might need to be adjusted. A plausible hypothesis is that immediate future outcomes, measured in minutes or hours, could correlate with brain dynamics over shorter time periods. The validity of this hypothesis in the context of other datasets remains an area for future investigation.

\section{Conclusion}

In this paper, we introduced a Graph Neural Network with Transformer Fusion (GNN-TF) framework designed to integrate brain dynamics with tabular covariates for predictive purposes. Based on the extensive experimental comparisons conducted using the NCANDA dataset, this approach demonstrated superior performance in comparison to other existing methods. Given its efficacy, this methodological proposal holds potential utility for other functional brain imaging studies where the integration of multiple data types is crucial for predicting clinical outcomes. The versatility and effectiveness of this approach in integrating diverse data types make it a promising tool for advancing research in the field of brain imaging and its clinical applications.


    \section*{Acknowledgments}
    We thank the NCANDA study and its investigators for providing the raw data analyzed in this project.

    \section*{Funding}
    This work was supported in part by the NIH grants R01MH126970, R01EB022911, RF1AG079324,  RF1AG074204, R01NS133743, and P30AI161943.
    The NCANDA data collection was supported by NIH grants AA021697, AA021697-04S1, AA021695, AA021692, AA021696, AA021681, AA021690, and AA021691.

    \section*{Ethical considerations}
    Only de-identified  data derived from the NCANDA study were provided to us for this secondary-analysis study, and we have no way to link the coded data to individual identities. Consequently, this study is classified as non–human-subjects research and does not require IRB approval.

    We are not affiliated with the NCANDA research team. The NCANDA research team obtained Institutional Review Board (IRB) approval at each participating site, and all data collection procedures were conducted in accordance with U.S. Department of Health and Human Services regulations (45 CFR Part 46).

    \section*{Data availability statement}
    The data that support the findings of this study are openly available in figshare  at \url{http://doi.org/10.6084/m9.figshare.29318804}, reference number\newline [10.6084/m9.figshare.29318804]. The code implementation of our research will be made publicly available on GitHub upon the acceptance of our work for publication.

    The raw data analyzed in this study were obtained from the National Consortium on Alcohol and Neurodevelopment in Adolescence (NCANDA). Access to the dataset is restricted to protect human subjects,  and was granted to us under a non-transferable, non-distributable data-use agreement. Researchers can obtain the raw data by completing  the data-use agreements with the NCANDA study and the  National Institute on Alcohol Abuse and Alcoholism (NIAAA),  following the application procedure described at \url{http://www.ncanda.org/datasharing.php}.


    \section*{Author contribution statement}

    RZ contributed to conceptualization, formal analysis, investigation, methodology, software, validation, visualization, and writing—original draft of the manuscript. XL contributed to conceptualization, data curation, funding acquisition, methodology, resources, supervision, and writing—original draft, as well as review and editing of the manuscript.

    \section*{Conflict of interest}

    The authors declare no potential conflict of interests.


\bibliographystyle{apalike}
\bibliography{rossi}

\clearpage           
\clearpage

\end{document}